\documentclass[final]{cvpr}

\usepackage{dsfont}
\usepackage{times}
\usepackage{epsfig}
\usepackage{graphicx}
\usepackage{amsmath}
\usepackage{amssymb}
\usepackage{color} 
\usepackage[ruled]{algorithm2e}
\usepackage{algorithmic}
\usepackage{multirow}
\usepackage{subcaption}
\usepackage{tablefootnote}
\usepackage[labelsep=period]{caption}
\usepackage[pagebackref=true,breaklinks=true,colorlinks,bookmarks=false]{hyperref}

\def\cvprPaperID{4071} 
\def\confYear{CVPR 2021}

\begin{document}

\title{FixBi: Bridging Domain Spaces for Unsupervised Domain Adaptation}

\author{Jaemin Na$^1$, Heechul Jung$^2$, Hyung Jin Chang$^3$, and Wonjun Hwang$^1$\\
$^1$Ajou University, Korea, $^2$Kyungpook National University, Korea, $^3$University of Birmingham, UK\\
{\tt\small osial46@ajou.ac.kr, heechul@knu.ac.kr, h.j.chang@bham.ac.uk, wjhwang@ajou.ac.kr}
}

\maketitle

\begin{abstract}
Unsupervised domain adaptation (UDA) methods for learning domain invariant representations have achieved remarkable progress. However, most of the studies were based on direct adaptation from the source domain to the target domain and have suffered from large domain discrepancies. In this paper, we propose a UDA method that effectively handles such large domain discrepancies. We introduce a fixed ratio-based mixup to augment multiple intermediate domains between the source and target domain. From the augmented-domains, we train the source-dominant model and the target-dominant model that have complementary characteristics. Using our confidence-based learning methodologies, e.g., bidirectional matching with high-confidence predictions and self-penalization using low-confidence predictions, the models can learn from each other or from its own results. Through our proposed methods, the models gradually transfer domain knowledge from the source to the target domain. Extensive experiments demonstrate the superiority of our proposed method on three public benchmarks: Office-31, Office-Home, and VisDA-2017.
\footnote{
Our code is available at https://github.com/NaJaeMin92/FixBi.}
\end{abstract}

\section{Introduction}
Recently, we have seen considerable improvements in several computer vision applications using deep learning; however, this success has been limited to supervised learning methods with abundant labeled data. 
Collecting and labeling data from various domains is an expensive and time-consuming task. To address this problem, semi-supervised learning~\cite{S4L, MixMatch, FixMatch} and unsupervised learning~\cite{Gidaris2018} have been studied; however, in most cases, it was assumed that learning of the model occurred in a similar domain.

UDA refers to a set of transfer learning methods for transferring knowledge learned from the source domain to the target domain under the assumption of domain discrepancy. Moreover, it is useful when the source domain contains enough labeled data to learn, but not much labeled data are present in the target domain. Domain adaptation (DA) generally assumes that the two domains have the same conditional distribution, but different marginal distributions. Under these assumptions, effective knowledge transfer is difficult when the two domains have large marginal distribution gaps. This becomes much more challenging in a scenario where the target domain has no labeled data at all.

\begin{figure}[t]
\centering
\includegraphics[width=0.95\columnwidth]{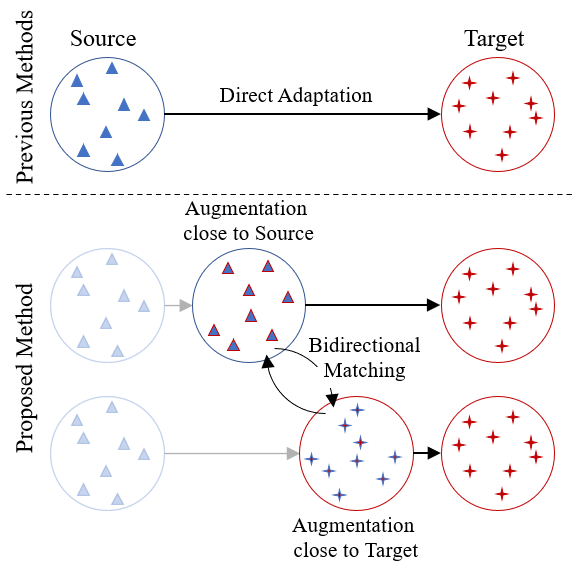}
\caption{Comparison of previous domain adaptation methods and our proposed method. \textbf{Top:} Previous methods try to adapt directly without any consideration of large domain discrepancies. \textbf{Bottom:} 
Our proposed method utilize augmented domains between the source and target domain for efficient domain adaptation.
}
\label{fig:01}
\end{figure}

In previous UDA methods, a domain discriminator~\cite{Ganin2015, ADDA2017} was introduced to encourage domain confusion through domain-adversarial objectives and minimize the gap between the source and target distributions. In~\cite{Long2015, Long2017}, domain discrepancy based approaches used metrics such as maximum mean discrepancy (MMD) and joint MMD (JMMD) to reduce the difference between two feature spaces. Moreover, inspired by the generative adversarial network (GAN), GAN-based DA methods~\cite{CyCADA2017, Shuhao2020} have attempted to generate transferable representations to minimize domain discrepancy. Most of these studies have directly adapted the knowledge learned from the source domain to the target domain. However, fundamentally, this does not take into account the case where the distance between the source and target domain is large, as shown in Figure~\ref{fig:01}.

In this paper, our goal is to compensate efficiently for the large domain discrepancies. To address this challenge, we construct multiple intermediate augmented domains, whose characteristics are different and complementary to each other. To achieve this, we propose a fixed ratio-based mixup. Our proposed mixup approach minimizes the domain randomness of~\cite{Wu2020, Minghao2020} between the source and target samples and generates multiple intermediate domains, as shown in Figure~\ref{fig:01}. For example, an augmented domain close to the source domain has more reliable label information, but it has a lower correlation with the target domain. By contrast, label information in an augmented domain close to the target domain is relatively inaccurate, but the similarity to the target domain is much higher. 

In these augmented domains, we train the complementary models that teach each other to bridge between the source and target domain. Specifically, we introduce a bidirectional matching based on the high-confidence predictions of each model for the target samples, moving the intermediate domains to the target domain. We also apply self-penalization, which penalizes its own model to improve performance through self-training. Moreover, to properly impose the characteristics of models that change with each iteration, we use an adaptive threshold by the confidence distribution of each mini-batch, not a predefined one~\cite{Co-teaching, MiCo2020}. Finally, to prevent divergence of the augmented models generated in different domains, we propose a consistency regularization using an augmented domain with the same ratio of source and target samples.

We conduct extensive ablation studies for a detailed analysis of our proposed method and achieve comparable performance to state-of-the-art methods in standard DA benchmarks such as Office-31~\cite{Office-31}, Office-Home~\cite{Office-HOME}, and VisDA-2017~\cite{VisDA2017}. The main contributions of this paper are summarized as follows.
\begin{itemize}
    \item We propose a fixed ratio-based mixup to efficiently bridges the source and target domains utilizing the intermediate domains.
    \item We propose confidence-based learning methodologies: a bidirectional matching and a self-penalization using positive and negative pseudo-labels, respectively.
    \item We empirically validate the superiority of our method to UDA with extensive ablation studies and evaluations on three standard benchmarks.
\end{itemize}


\section{Related Work}
\textbf{Semi-supervised Learning.} Semi-supervised learning (SSL)~\cite{ReMixMatch, MixMatch, DLEE2013, Consistency2016, S4L, FixMatch, NS3L} leverages unlabeled data to improve a model’s performance when limited labeled data is provided, which alleviates the expensive labeling process efficiently. Many recently proposed semi-supervised learning methods, such as MixMatch~\cite{MixMatch}, FixMatch~\cite{FixMatch}, and ReMixMatch~\cite{ReMixMatch}, based on augmentation viewpoints. MixMatch~\cite{MixMatch} used low-entropy labels for data-augmented unlabeled instances and mixed labeled and unlabeled data for semi-supervised learning. 
On the basis of consistency regularization and pseudo-labeling, FixMatch~\cite{FixMatch} generates pseudo-labels using the model’s predictions on weakly augmented unlabeled images. Then, when the examples have high-confidence predictions, they train the model using strong-augmented images. Note that in general, they assumed that labeled and unlabeled data have similar domains or feature distributions. 

Basically, semi-supervised domain adaptation has more information about some target labels compared with UDA, and some related works~\cite{Shuang2017, Kuniaki2019, Qin2020, Dali2020, MiCo2020} have been proposed leveraging semi-supervised signals. Specifically, in~\cite{Kuniaki2019}, a minimax entropy approach was proposed that adversarially optimizes an adaptive few-shot model. In~\cite{Qin2020}, the learning of opposite structures was unified whereby it consists of a generator and two classifiers trained with opposite forms of losses for a unified object. 

Meanwhile,~\cite{MiCo2020} addresses semi-supervised domain adaptation by breaking it down into SSL and UDA problems. Two models are in charge of each sub-problem and are trained based on co-teaching. One model is trained with labeled source samples and labeled target samples, and the other model is trained with unlabeled target samples and labeled target samples. In this way, by using different combinations of data, it provides two different perspectives. By contrast, in this paper, we guarantee two different perspectives with the two types of our fixed ratio-based mixup. In addition, \cite{MiCo2020} used co-teaching~\cite{Co-teaching} concepts to train the mixup objectives between the source and target domain, whereas we use this concept to train the pseudo-labels in the target domain with bidirectional matching. Furthermore, when applying the mixup operation,~\cite{MiCo2020} uses only selected target samples, whereas we use all target samples.

\begin{figure*}[th]
\centering
\includegraphics[width=1.0\linewidth]{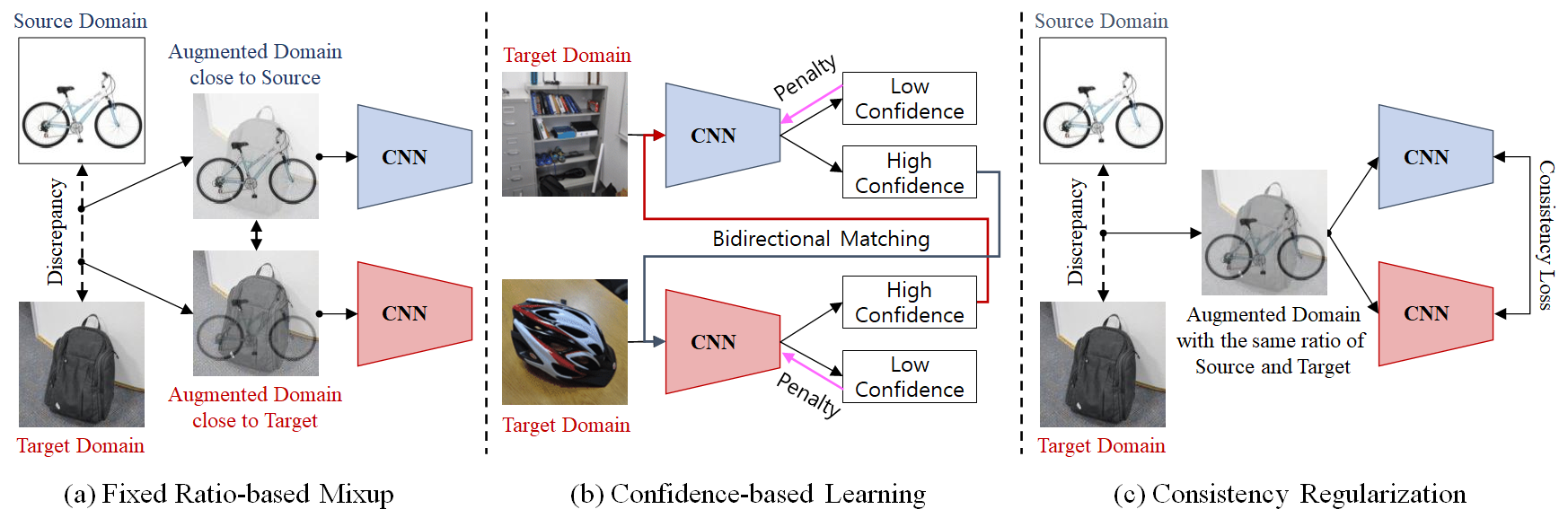}
\caption{\textbf{An overview of the proposed method.}
The proposed method consists of (a) fixed ratio-based mixup, (b) confidence-based learning, e.g., bidirectional matching with the positive pseudo-labels and self-penalization with the negative pseudo-labels, and (c) consistency regularization. Best viewed in color.}
\label{fig:02}
\end{figure*}

\textbf{Unsupervised Domain Adaptation.}
Recent works~\cite{Gopalan2011, SUN2016, Tzeng2017, Long2017, Ganin2015, ADDA2017, Saito2018, Minghao2020} have focused on UDA based on domain alignment and discriminative domain-invariant feature learning methods. 
For example, a deep adaptation network (DAN)~\cite{Long2015} minimized MMD over domain-specific layers, and joint adaptation networks \cite{Long2017} aligned the joint distributions of domain-specific layers across different domains based on a JMMD. Deep domain confusion (DCC)~\cite{Tzeng2017} made use of MMD metrics in the fully connected layer for learning both discriminative and transferable domains.
A domain adversarial neural network (DANN)~\cite{Ganin2015} learned a domain invariant representation by back-propagating the reverse gradients of the domain classifier. Adversarial discriminative domain adaptation (ADDA)~\cite{ADDA2017} learned a discriminative representation using the source labels, and then, a separate encoding that maps the target data to the same space based on a domain-adversarial loss is used. Maximum classification discrepancy (MCD) \cite{MCD2018} tried to align the distribution of a target domain by considering task-specific decision boundaries by maximizing the discrepancy on the target samples and then generating features that minimize this discrepancy. 
A contrastive adaptation network (CAN)~\cite{CAN2019} optimized the metric for minimizing the domain discrepancy, which explicitly models the intra-class domain discrepancy and the inter-class domain discrepancy. 

Robust spherical domain adaptation (RSDA)~\cite{Gu2020} used a spherical classifier for label prediction and a spherical domain discriminator for discriminating domain labels and utilized robust pseudo-label loss in the spherical feature space. Structurally regularized deep clustering (SRDC)~\cite{SRDC} enhanced target discrimination by clustering intermediate network features and structural regularization with soft selection of less divergent source examples. 
Dual mixup regularized learning (DMRL)~\cite{Wu2020} guided the classifier to enhance consistent predictions between samples and enriched the intrinsic structures of the latent space. For mixing the source and target domains, they proposed two mixup regularizations based on randomness. 

Note that in this study, we create bridges between the target and source domain by augmenting multiple intermediate domains. For this purpose, unlike~\cite{Wu2020, Minghao2020}, the scope of the augmented domain was not expanded simply by relying on randomness. However, two fixed ratio-based mixups are used to create a source-closed augmented domain, which has a clear label but is at a distance from the target domain, and a target-closed augmented domain, which has the opposite properties. Then, they teach each other in order to transfer domain knowledge to the target side.

\section{Proposed Method}
In UDA, we are given labeled data $\mathcal{X}^s = \{({x}^{s}_i, {y}^{s}_i)\}^{N_s}_{i=1}$ from the source domain and unlabeled data $\mathcal{X}^t = \{({x}^{t}_i)\}^{N_t}_{i=1}$ from the target domain where the $N_{s}$ and $N_{t}$ denote the sizes of $\mathcal{X}^s$ and $\mathcal{X}^t$, respectively. The large distribution gap between $P(\mathcal{X}^s)$ and $P(\mathcal{X}^t)$ is one of the major obstacles for the UDA problem. Our goal is to ensure that the knowledge learned from the source domain is well generalized in the target domain. 
In this section, we introduce our FixBi algorithm and the detailed ideas it builds on, as shown in Figure~\ref{fig:02}. 

\subsection{Fixed Ratio-based Mixup}
In general, the mixup~\cite{MixUp} is a kind of data augmentation method to increase the robustness of neural networks when learning from corrupt labels. Recent studies \cite{MixMatch, FixMatch} have utilized the 
mixup to construct virtual samples with convex combinations between labeled and unlabeled data. 
In this context, most domain adaptation methods \cite{VMT, Minghao2020, MiCo2020, Wu2020} based on the mixup use mixup-ratio $\lambda$ with randomly sampled values from the beta distribution: $\lambda \sim Beta(\alpha, \alpha)$ where $\alpha$ is a hyperparameter.
It is because they have tried to generate training samples that exist somewhere between the source and target domain without any consideration of the domain gap. However, we propose to use two fixed mixup ratios $\lambda_{sd}$ and $\lambda_{td}$ to provide more clarity and less randomness.

Given a pair of input samples and their corresponding one-hot labels in the source and target domain: ($x^s_i, y^s_i$) and ($x^t_i, \hat{y}^{t}_i$), our mixup settings are defined as follows:
\begin{equation}
\begin{gathered}
    \Tilde{x}^{st}_i = \lambda {x}^{s}_i +(1-\lambda) {x}^{t}_i \\
    \Tilde{y}^{st}_i = \lambda {y}^{s}_i +(1-\lambda) \hat{y}^{t}_i,
\end{gathered}
\end{equation}

where $\lambda \in \{\lambda_{sd}, \lambda_{td}\}\; s.t. \; \lambda_{sd} + \lambda_{td} = 1$. Note that $\hat{y}^{t}_i$ is the pseudo-labels obtained by the baseline model, e.g., DANN~\cite{Ganin2015} or MSTN~\cite{MSTN}, for the unlabeled target samples. The detailed analysis of our fixed ratio-based mixup is covered in Section 4.2.

Taking advantage of the fixed ratio-based mixup, we construct two network models that act as bridges between the source and target domain. The key point here is to obtain two networks with different perspectives through our mixup strategies. For this purpose, we leverage two different models made by the proposed mixup ratios $\lambda_{sd}$ and $\lambda_{td}$: \emph{“source-dominant model”} (SDM) and \emph{“target-dominant model”} (TDM). The source-dominant model has strong supervision for the source domain but relatively weak supervision for the target domain. By contrast, the target-dominant model has strong target supervision but weak source supervision. As both types of mixups are not confined to a single domain, they can serve as bridges between the two different domains.

Consequently, we apply two fixed ratios $\lambda_{sd}$ for SDM and $\lambda_{td}$ for TDM. Let $p(y|\Tilde{x}_i^{st})$ denote the predicted class distribution produced by the model for an input $\Tilde{x}_i^{st}$. Then the objective of our fixed ratio-based mixup is defined as follows:
\begin{equation}
    \mathcal{L}_{fm}= \frac{1}{B}\sum_{i=1}^{B}{\hat{y}_i^{st} \log(p(y|\Tilde{x}_i^{st}))},
\end{equation}
where $\hat{y}^{st}_i=\operatorname*{argmax}p(y|\Tilde{x}^{st}_i)$ and B is a mini-batch size.

\subsection{Confidence-based Learning}
Through our fixed ratio-based mixup, the two networks have different characteristics and can develop with mutually complementary learning. To utilize the two models as bridges from the source domain to the target domain, we propose a confidence-based learning where one model teaches the other model using the positive pseudo-labels or teach itself using the negative pseudo-labels.

\textbf{Bidirectional Matching with positive pseudo-labels.} 
Inspired by \cite{FixMatch, Co-teaching, Co-training}, when one network assigns the class probability of input above a certain threshold $\tau$, we assume that this predicted label as a pseudo-label. Here, we refer to these labels as positive pseudo-labels. Then we train the peer network to make its predictions match these positive pseudo-labels via a standard cross-entropy loss. Let us denote probability distributions $p$ and $q$ of two models. Then the objective of our bidirectional matching is defined as follows:
\begin{equation}
\begin{gathered}
    \mathcal{L}_{bim} = \frac{1}{B}\sum_{i=1}^{B}{\mathds{1}(max(p(y|{x}_i^{t})>\tau) \hat{y}_i^{t} \log(q(y|x_i^{t}))},
\end{gathered}
\end{equation}
where $\hat{y}^{t}_i = \operatorname*{argmax}p(y|{x}_i^{t})$. Note that in \cite{FixMatch}, only one-way matching is used according to input augmentations. However, since our method derives the results from both networks for the same input, bidirectional matching is available.

\textbf{Self-penalization with negative pseudo-labels.} As well as the bidirectional matching that matches the positive pseudo-labels to the predictions of the peer network, each network learns through the self-penalization using the negative pseudo-labels.
Here, the negative pseudo-label indicates the most confident label (\emph{top-1} label) predicted by the network with a confidence lower than the threshold $\tau$. Since the negative pseudo-label is unlikely to be a correct label, we need to increase the probability values of all other classes except for this negative pseudo-label. Therefore, we optimize the output probability corresponding to the negative pseudo-label to be close to zero. The objective of self-penalization is defined as follows:
\begin{equation}
    \mathcal{L}_{sp} = \frac{1}{B}\sum_{i=1}^{B}{\mathds{1}(max(p(y|{x}_i^{t})<\tau) \hat{y}_i^{t} \log(1-p(y|x_i^{t}))}.
\end{equation}

Unlike the recent studies \cite{Co-teaching, FixMatch, MiCo2020} that ignore the low-confidence predictions (or large loss samples), it is worth noting that we utilize the low-confidence predictions as the meaningful knowledges for learning the models. Furthermore, we apply the learnable temperature of the softmax to adjust the output distributions.

Looking back to the confidence threshold $\tau$, the basic strategy is to set a fixed value as a hyperparameter.
However, a deep neural network (DNN) tends to start at a low level of confidence value and its value gradually increases as the network learns. The fixed threshold cannot properly reflect the confidence which is changed constantly during training, therefore the number of positive and negative pseudo-labels can be biased to one side. To overcome this problem, we adopt an adaptive threshold which is changed adaptively by the sample mean and standard deviation of a mini-batch, not a fixed one.

\subsection{Consistency Regularization}
Through our confidence-based learning, the two models with different characteristics gradually get closer to the target domain because they are trained with more reliable pseudo-labels of the target samples. We introduced a new consistency regularization to ensure a stable convergence of training both models. Here, we assume that the well-trained models should be regularized to have consistent results in the same space. It helps to construct the domain bridging by ensuring that the two models trained from the different domain spaces maintain consistency in the same area between the source and target domain. For the intermediate space, both fixed mixup-ratios $\lambda_{sd}$ and $\lambda_{td}$ are set to 0.5. The consistency regularization loss can be defined as follows:
\begin{equation}
     \mathcal{L}_{cr} = \frac{1}{B}\sum_{i=1}^{B}\Vert p(y|\Tilde{x}_i^{st}) - q(y|\Tilde{x}_i^{st}) \Vert^2_2.
\end{equation}

\begin{algorithm}
\caption{FixBi Training Procedure}
\label{algo1}
\SetAlgoLined
\SetKwInOut{Input}{Input}
\SetKwInOut{Output}{Output}
\Input{
    Network weights $\textcolor{blue}{w_{sd}}$ and $\textcolor{red}{w_{td}}$, total epochs $E$, mini batch $B$, 
    warm-up epochs $k$, mixup-ratios $\lambda_{sd}$, $\lambda_{td}$, and $\lambda_{cr}(=0.5)$,
    source samples ${x}^s$, target samples ${x}^t$, and mixup samples $\Tilde{M}$.
}
 \For{$e$=1 \KwTo $E$}{
    \For{$i$=1 \KwTo $B$}{
        \textbf{Obtain} $\Tilde{M}_{sd}$ using Eq. (1) with $\lambda_{sd}$;\\
        \textbf{Obtain} $\Tilde{M}_{td}$ using Eq. (1) with $\lambda_{td}$;\\
        \textbf{Update} $\mathcal{L}_{fm}(\Tilde{M}_{sd}; \textcolor{blue}{w_{sd}})$ and $\mathcal{L}_{sp}({x}^{t}; \textcolor{blue}{w_{sd}});$\\
        \textbf{Update} $\mathcal{L}_{fm}(\Tilde{M}_{td}; \textcolor{red}{w_{td}})$ and $\mathcal{L}_{sp}({x}^{t}; \textcolor{red}{w_{td}});$\\
        \If{$e > k$}{
            \If{$max(y|x^t; \textcolor{red}{w_{td}}) > \tau_{td}$}{
                \textbf{Update} $\mathcal{L}_{bim}({x}^{t}; \textcolor{blue}{w_{sd}});$\\
            }
            \If{$max(y|x^t; \textcolor{blue}{w_{sd}}) > \tau_{sd}$}{
                \textbf{Update} $\mathcal{L}_{bim}({x}^{t}; \textcolor{red}{w_{td}});$\\
            }
            \textbf{Obtain} $\Tilde{M}_{cr}$
            using Eq. (1) with $\lambda_{cr};$\\
            \textbf{Update}  $\mathcal{L}_{cr}(\Tilde{M}_{cr}; \textcolor{blue}{w_{sd}});$\\
            \textbf{Update}  $\mathcal{L}_{cr}(\Tilde{M}_{cr}; \textcolor{red}{w_{td}});$
        }
    }
 }
\Output{Learned model parameters $\textcolor{blue}{w_{sd}}$ and $\textcolor{red}{w_{td}}$.}
\end{algorithm}

\subsection{Training Procedure}
The training process of our FixBi is summarized in Algorithm \ref{algo1}. First, we start to train our networks with pre-trained baseline weights, similar to \cite{Gu2020}. Then, we copy the pre-trained weights to $w_{sd}$ and $w_{td}$. In each iteration, we generate two types of samples with different mixup ratios $\lambda_{sd}$ and $\lambda_{td}$. Initially, to ensure that the two networks have independent characteristics, we apply a warm-up period of $k$ epochs where each network is independently trained only with the fixed ratio-based mixup and the self-penalization. After enough training, we begin to train with bidirectional matching that can teach each other. Note that one network is trained with pseudo-labels from the peer network which satisfies the confidence threshold constraint. At the same time, we also apply the consistency regularization loss to guarantee stable convergence in training.

\begin{table}[t]
\centering
\caption{Comparison of three different mixup-ratio rules on the task A$\rightarrow$W.}
\label{table1}
\resizebox{0.99\linewidth}{!}{%
\begin{tabular}{c|c|c|cc}
\hline
\multirow{2}{*}{Type} & \multicolumn{2}{c|}{w/o $\mathcal{L}_{bim}$} & \multicolumn{2}{c}{w/ $\mathcal{L}_{bim}$}                       \\ \cline{2-5} 
                      & SDM          & TDM         & \multicolumn{1}{c|}{SDM}         & TDM         \\ \hline \hline
Random                & 86.5$\pm$1.0  & 85.3$\pm$0.9 & \multicolumn{1}{c|}{86.7$\pm$0.8} & 85.6$\pm$0.7 \\ \hline
Range                 & 86.0$\pm$1.7  & 29.6$\pm$6.8 & \multicolumn{1}{c|}{83.3$\pm$6.2} & 81.0$\pm$5.4 \\ \hline
Fixed (Ours)                & 86.3$\pm$0.6  & 86.0$\pm$0.7 & \multicolumn{1}{c|}{89.3$\pm$0.4} & 90.1$\pm$0.3 \\ \hline
\end{tabular}
}
\end{table}

\section{Experiments}
We evaluate our proposed method on three domain adaptation benchmarks such as Office-31, Office-Home and VisDA-2017, compared with state-of-the-art domain adaptation methods. In addition, we validate the contributions of the proposed method through extensive ablation studies.
\subsection{Setups}
\textbf{Datasets.} We evaluated our method in the following three standard benchmarks for UDA.

\textbf{Office-31} \cite{Office-31} is the most popular dataset for real-world domain adaptation. It contains 4,110 images of 31 categories in three domains: Amazon (A), Webcam (W), DSLR (D). We evaluated all methods on six domain adaptation tasks. 

\textbf{Office-Home} \cite{Office-HOME} is a more challenging benchmark than Office-31. It consists of images of everyday objects organized into four domains: artistic images (Ar), clip art (Cl), product images (Pr), and real-world images (Rw). It contains 15,500 images of 65 classes. 

\textbf{VisDA-2017} \cite{VisDA2017} is a large-scale dataset for synthetic-to-real domain adaptation. It contains 152,397 synthetic images for the source domain and 55,388 real-world images for the target domain.

\textbf{Baselines.} Since our proposed method can be flexibly applied in any UDA methods, we use DANN~\cite{Ganin2015} as a baseline for a detail analysis of our contributions and MSTN~\cite{MSTN} for performance comparisons with the state-of-the-art methods.

\textbf{Implementation details.} Following the standard protocol for UDA, we use all labeled source data and all unlabeled target data. For Office-31 and Office-Home, we use ResNet-50 \cite{ResNet1, ResNet2} as the backbone network. We use mini-batch stochastic gradient descent (SGD) with a momentum of 0.9, an initial learning rate of 0.001, and a weight decay of 0.005. We follow the same learning rate schedule as in \cite{Ganin2015}. The mixup ratios are set up with $\lambda_{sd}= 0.7$ and $\lambda_{td}= 0.3$. The confidence threshold $\tau$ is calculated as $({mean} - 2\times{std})$ across all mini-batches. We train the model for a total of 200 epochs and set the warm-up epochs to 100. For VisDA-2017, we use ResNet-101 as the backbone architecture. We use the SGD optimizer with a momentum of 0.9, an initial learning rate of 0.0001, and a weight decay of 0.005. We train the model for 25 epochs with warm-up epochs of 10. 

\subsection{Ablation studies and discussions}
For a more detailed analysis of our proposed method, we conducted ablation studies on the Office-31 dataset with DANN \cite{Ganin2015} as the baseline model. 

\begin{figure}[t]
\centering
\includegraphics[width=0.9\columnwidth]{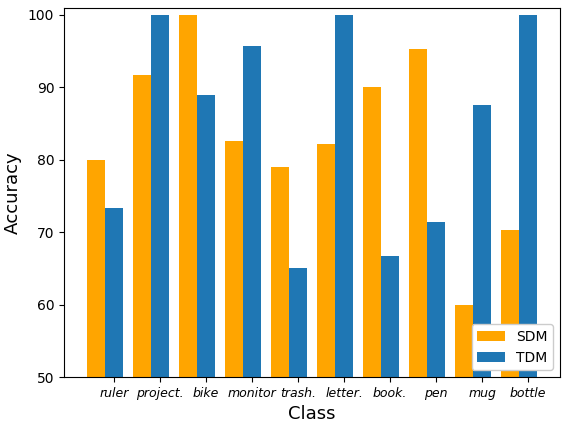}
\caption{Class-wise accuracy (\%) on the task A$\rightarrow$W of the Office-31. Best viewed in color.}
\label{fig:03}
\end{figure}

\textbf{Comparison of different mixup-ratio rules.} We compare our fixed-ratio mixup with two existing general ratios in Table~\ref{table1}. The \emph{“Random”} refers to the ratio randomly sampled from the beta distribution, as in the traditional mixup approaches \cite{MixUp, MiCo2020, VMT, Minghao2020}. The \emph{“Range”} refers to the ratio randomly sampled from the beta distribution and limited to a specific range. In this case, the mixup ratio $\lambda^{\prime}$ is determined by $\lambda^{\prime}\sim$ max$(\lambda, 1- \lambda)$, where $\lambda\sim$ Beta($\alpha,\alpha$). In this experiment, SDM and TDM are trained with $\lambda^{\prime}$ and $1-\lambda^{\prime}$, respectively, intending to give them different perspectives. We set the hyperparameter $\alpha$ of the beta distribution to 1.0 for~\emph{“Random”} and \emph{“Range”}, as used in \cite{MiCo2020, VMT}. Lastly, the \emph{“Fixed”} refers to using two fixed mixup ratios ($\lambda_{sd}$, $\lambda_{td}$). We set $\lambda_{sd}=0.7$ and $\lambda_{td}=0.3$, which satisfies $\lambda_{sd}+\lambda_{td}=1$. Note that our final accuracy is the result of an ensemble of the output probabilities of both models.

The left side of Table~\ref{table1} shows the accuracy when only a fixed ratio mixup is applied without the bidirectional matching. In the case of \emph{“Random”}, the accuracy is similar to that of \emph{“Fixed”}. However, in the case of \emph{“Range”}, extreme accuracy degradation is noticeable in TDM. It shows that a mixup which is too target-biased negatively affects learning when the target labels are incorrect. The right side of Table~\ref{table1} presents the accuracy when applying the bidirectional matching with the fixed ratio mixup. In the case of \emph{“Random”} and \emph{“Range”}, it is difficult to expect performance improvement through the bidirectional matching. By contrast, we observe that the networks trained through the fixed ratio-based mixup can benefit from the bidirectional matching.  

\textbf{Why is it better to use a fixed ratio?} We claim that this difference occurs because the two networks have different perspectives through our fixed ratio-based mixup. To verify this, we analyzed the class-wise accuracy of SDM and TDM. We apply $\mathcal{L}_{fm}$ with fixed ratios $\lambda_{sd}=0.7$ and $\lambda_{td}=0.3$ to the models, respectively. We pick the top-10 accuracies with the largest difference between class-wise accuracies for the two network models. As shown in Figure~\ref{fig:03}, we observe that these two models have different strengths and weaknesses from the viewpoint of class-wise accuracy. Note that the performances of these two models are similar at 86.3\% and 86.0\%.

\begin{figure}[t]
\hspace{-3.0mm}
\centering
\includegraphics[width=1.0\columnwidth]{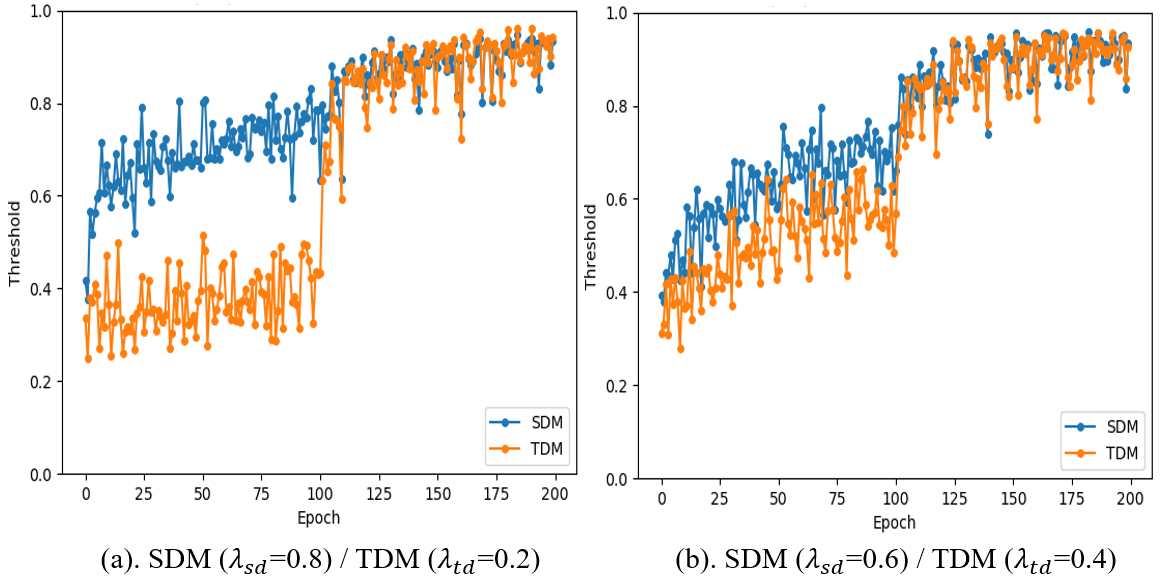}
\caption{Visualization of the confidence threshold $\tau$ on the task A$\rightarrow$W of the Office-31. Best viewed in color.}
\label{fig:04}
\end{figure}

\begin{figure}[t]
    \hspace{1.5mm}
    \centering
    \begin{subfigure}[b]{0.45\columnwidth}
        \includegraphics[width=\columnwidth]{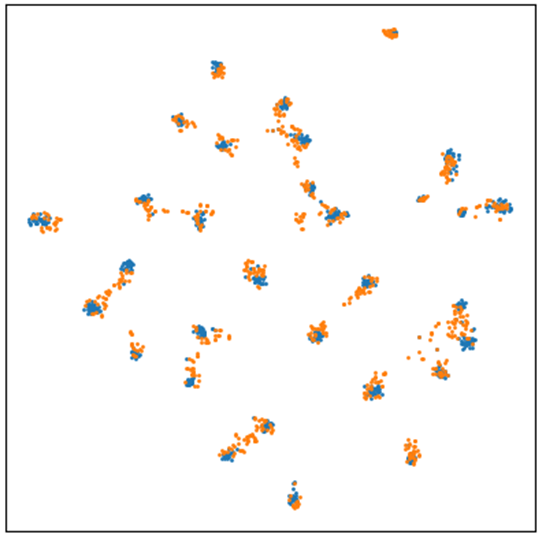}
        \caption{Baseline (DANN)}
        \label{fig:05_1}
    \end{subfigure}
    \begin{subfigure}[b]{0.45\columnwidth}
        \includegraphics[width=\columnwidth]{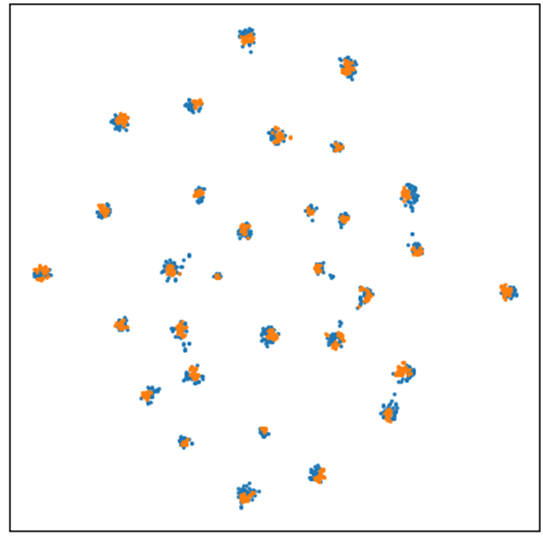}
        \caption{FixBi (Ours)}
        \label{fig:05_2}
    \end{subfigure}
    \vspace{1.5mm}
    \caption{The visualization of embedded features on the task A$\rightarrow$W. Blue and orange points denote the source and target domains, respectively. Best viewed in color.}
    \label{fig:05}
\end{figure}

\textbf{Comparison with simple ensemble models.} To show that our two perspectives have a different meaning to a simple ensemble of a single-perspective, we compare the ensemble of a single-perspective with our proposed method. For the single-perspective, we train models with the mixup ratios of (0.3, 0.3) and (0.7, 0.7), respectively. For a fair comparison, we apply only $\mathcal{L}_{fm}$ and $\mathcal{L}_{bim}$. As shown in Table~\ref{table2}, the accuracy of our two-perspective model is higher than that of the other single-perspective models.

\begin{table*}[t]
\centering
\caption{Comparison of ensemble networks on Office-31.}
\label{table2}
\begin{tabular}{c|c|c|c|c|c|c|c|c}
\hline
      Method   & ($\lambda_{sd}$, $\lambda_{td}$) & A$\rightarrow$W & D$\rightarrow$W & W$\rightarrow$D & A$\rightarrow$D & D$\rightarrow$A & W$\rightarrow$A & Avg           \\ \hline \hline
\multirow{2}{*}{Single-perspective} & (0.3, 0.3) & 88.6 & 96.5 &100.0 & 85.6 & 69.4 & 65.1 & 84.2 \\ 
\cline{2-9}
                   & (0.7, 0.7) & 89.2 & 96.5 &100.0 & 85.5 & 69.1 & 67.8 & 84.7 \\ \hline
Two-perspective (Ours) & (0.7, 0.3)                       & \textbf{90.1}   & \textbf{98.5}   & \textbf{100.0} & \textbf{88.4}   & \textbf{72.5}   & \textbf{72.5}   & \textbf{87.0} \\ \hline
\end{tabular}
\end{table*}

\begin{table*}[t]
\begin{center}
\caption{Ablation results (\%) of investigating the effects of our components on Office-31.}
\label{table3}
\begin{tabular}{c|c|c|c|c||c|c|c|c|c|c|c c}
\hline{}
$\mathcal{L}_{DANN}$ & $\mathcal{L}_{fm}$ & $\mathcal{L}_{bim}$ & $\mathcal{L}_{sp}$ & $\mathcal{L}_{cr}$ & A$\rightarrow$W & D$\rightarrow$W & W$\rightarrow$D & A$\rightarrow$D & D$\rightarrow$A &W$\rightarrow$A & Avg\\
\hline \hline
\checkmark& &	 &	 &	& 82.0&	96.9&	99.1&	79.7&	68.2&	67.4& 82.2\\
\checkmark&	\checkmark&	 && 	&86.5&	98.4&	\textbf{100.0}&	85.5&	71.4&	71.5& 85.5\\
\checkmark& \checkmark&	\checkmark&	 &	& 90.1&	98.5&	\textbf{100.0}&	88.4&	72.5&	72.5& 87.0\\
\checkmark& \checkmark&	\checkmark& \checkmark&	& 92.3&	98.6&	\textbf{100.0}&	90.4&	76.3& 74.1&	88.6\\
\checkmark& \checkmark&	\checkmark& \checkmark&	\checkmark& \textbf{94.2}&	\textbf{99.3}&	\textbf{100.0}&	\textbf{91.3}&	\textbf{76.5}& \textbf{74.3}&	\textbf{89.3}\\
\hline
\end{tabular}
\end{center}
\end{table*}

\begin{table*}[h!]
\begin{center}
\caption{Accuracy (\%) on Office-31 for unsupervised domain adaptation (ResNet-50). The best accuracy is indicated in bold and the second best one is underlined. * Reproduced by \cite{DSBN}}
\label{tab:Office-31}
\begin{tabular}{ccccccc|c}
\hline
Method       & A$\rightarrow$W & D$\rightarrow$W & W$\rightarrow$D & A$\rightarrow$D & D$\rightarrow$A & W$\rightarrow$A                    & Avg          \\ 
\hline\hline
ResNet-50~\cite{ResNet1} & 68.4±0.2          & 96.7±0.1       & 99.3±0.1           & 68.9±0.2          & 62.5±0.3 & \multicolumn{1}{c|}{60.7±0.3} & 76.1 \\
DANN~\cite{Ganin2015}   & 82.0±0.4          & 96.9±0.2       & 99.1±0.1           & 79.7±0.4          & 68.2±0.4 & \multicolumn{1}{c|}{67.4±0.5} & 82.2 \\
MSTN*~\cite{MSTN}      & 91.3              & 98.9           & \textbf{100.0}     & 90.4              & 72.7     & \multicolumn{1}{c|}{65.6}     & 86.5 \\
CDAN+E~\cite{CDAN}      & 94.1±0.1          & 98.6±0.1       & \textbf{100.0±0.0} & 92.9±0.2          & 71.0±0.3 & \multicolumn{1}{c|}{69.3±0.3} & 87.7 \\
DMRL~\cite{Wu2020}      & 90.8±0.3          & 99.0±0.2       & \textbf{100.0±0.0} & 93.4±0.5          & 73.0±0.3 & \multicolumn{1}{c|}{71.2±0.3} & 87.9 \\
SymNets~\cite{SymNets}   & 90.8±0.1          & 98.8±0.3       & \textbf{100.0±0.0} & 93.9±0.5          & 74.6±0.6 & \multicolumn{1}{c|}{72.5±0.5} & 88.4 \\
GSDA~\cite{GSDA}      & 95.7              & 99.1           & \textbf{100}       & 94.8              & 73.5     & \multicolumn{1}{c|}{74.9}     & 89.7 \\
CAN~\cite{CAN2019}       & 94.5±0.3          & 99.1±0.2       & \underline{99.8±0.2}           & 95.0±0.3          & \underline{78.0±0.3} & \multicolumn{1}{c|}{77.0±0.3} & 90.6 \\
SRDC~\cite{SRDC}      & \underline{95.7±0.2}          & \underline{99.2±0.1}       & \textbf{100.0±0.0}     & \textbf{95.8±0.2}          & 76.7±0.3 & \multicolumn{1}{c|}{77.1±0.1} & 90.8 \\
RSDA-MSTN~\cite{Gu2020} & \textbf{96.1±0.2} & \textbf{99.3±0.2}       & \textbf{100.0±0.0}       & \underline{95.8±0.3}          & 77.4±0.8 & \multicolumn{1}{c|}{\underline{78.9±0.3}} & \underline{91.1} \\ \hline
FixBi (Ours) &\textbf{96.1±0.2}            & \textbf{99.3±0.2}            & \textbf{100.0±0.0}  & 95.0±0.4            & \textbf{78.7±0.5}   & \multicolumn{1}{c|}{{\textbf{79.4±0.3}}} & \textbf{91.4} \\ \hline
\end{tabular}
\end{center}
\end{table*}

\textbf{Effects of the components of our FixBi.} We conduct ablation studies to investigate the effectiveness of the components of our proposed method. In Table~\ref{table3}, our fixed ratio-based mixup improves the baseline DANN \cite{Ganin2015} on average by 3.3\%. The bidirectional matching provides an additional 1.5\% improvement on average. Especially, in the task of A$\rightarrow$W and A$\rightarrow$D, we observe that the bidirectional matching has an impressive impact on performance improvement. We also observe that self-penalization has a significant impact on both task D$\rightarrow$A and W$\rightarrow$A. In addition, our consistency regularization loss helps to improve performance. Overall, FixBi improves the baseline DANN by an average of 7.1\%. This shows that each component is effective in improving performance.

\textbf{Analysis of the confidence threshold.} We visualize our confidence threshold $\tau$ in Figure~\ref{fig:04}. Note that our confidence threshold $\tau$ is changed adaptively within each mini-batch, and gradually increased with learning iterations. We observe a sharp increase in the confidence of TDM at 100 epoch, the point at which $\mathcal {L} _ {bim} $ starts to be applied. It can be seen more clearly, especially when the difference in the mixup ratio between SDM and TDM is large.

\textbf{Feature visualization.} Figure~\ref{fig:05} visualizes the embedded features on the task A$\rightarrow$W by t-SNE \cite{tSNE}. For the baseline (e.g, DANN \cite{Ganin2015}), the target domain features are embedded around the clusters of the source domain features, but it fails to form the clusters of the target domain features. On the other hand, our FixBi constructs the compact clusters of the target domain features close to the source domain features. From this result, we confirm that our proposed method works successfully for an unsupervised domain adaptation task.

\begin{table*}[t]
\caption{Accuracy (\%) on Office-Home for unsupervised domain adaptation (ResNet-50). The best accuracy is indicated in bold and the second best one is underlined. * Reproduced by \cite{Gu2020}}
\label{tab:Office-Home}
\resizebox{\textwidth}{!}{%
\begin{tabular}{ccccccccccccc|c}
\hline
Method &
  Ar$\rightarrow$Cl &
  Ar$\rightarrow$Pr &
  Ar$\rightarrow$Rw &
  Cl$\rightarrow$Ar &
  Cl$\rightarrow$Pr &
  Cl$\rightarrow$Rw &
  Pr$\rightarrow$Ar &
  Pr$\rightarrow$Cl &
  Pr$\rightarrow$Rw &
  Rw$\rightarrow$Ar &
  Rw$\rightarrow$Cl &
  Rw$\rightarrow$Pr &
  Avg \\ 
  \hline\hline
ResNet-50~\cite{ResNet1} & 34.9 & 50            & 58            & 37.4          & 41.9 & 46.2 & 38.5          & 31.2 & 60.4        & 53.9 & 41.2 & 59.9 & 46.1 \\
DANN~\cite{Ganin2015}   & 45.6 & 59.3          & 70.1          & 47            & 58.5 & 60.9 & 46.1          & 43.7 & 68.5        & 63.2 & 51.8 & 76.8 & 57.6 \\
CDAN~\cite{CDAN}      & 49   & 69.3          & 74.5          & 54.4          & 66   & 68.4 & 55.6          & 48.3 & 75.9        & 68.4 & 55.4 & 80.5 & 63.8 \\
MSTN*~\cite{MSTN}      & 49.8 & 70.3          & 76.3 & 60.4          & 68.5 & 69.6 & 61.4          & 48.9 & 75.7        & 70.9 & 55   & 81.1 & 65.7 \\
SymNets~\cite{SymNets}   & 47.7 & 72.9 & 78.5          & 64.2          & 71.3 & 74.2 & 63.6          & 47.6 & 79.4        & 73.8 & 50.8 & 82.6 & 67.2 \\
GSDA~\cite{GSDA}      & \textbf{61.3} & 76.1          & 79.4 & 65.4          & 73.3 & 74.3 & 65            & 53.2 & 80          & 72.2 & \underline{60.6} & 83.1 & 70.3 \\
GVB-GD~\cite{Shuhao2020}    & 57   & 74.7          & 79.8 & 64.6          & 74.1 & 74.6 & 65.2          & \underline{55.1} & 81          & 74.6 & 59.7 & 84.3 & 70.4 \\
RSDA-MSTN~\cite{Gu2020} & 53.2 & \textbf{77.7}          & \textbf{81.3} & 66.4          & 74   & 76.5 & \underline{67.9}          & 53   & \textbf{82} & 75.8 & 57.8 & \underline{85.4} & 70.9 \\
SRDC~\cite{SRDC}      & 52.3 & 76.3          & \underline{81}            & \textbf{69.5} & \underline{76.2} & \underline{78}   & \textbf{68.7} & 53.8 & \underline{81.7}        & \underline{76.3} & 57.1 & 85   & \underline{71.3} \\
\hline
FixBi (Ours) &
  \underline{58.1} &
  \underline{77.3} &
  80.4 &
  \underline{67.7} &
  \textbf{79.5} &
  \textbf{78.1} &
  65.8 &
  \textbf{57.9} &
  \underline{81.7} &
  \textbf{76.4} &
  \textbf{62.9} &
  \textbf{86.7} &
  \textbf{72.7} \\ \hline
\end{tabular}%
}
\end{table*}

\begin{table*}[t]
\caption{Accuracy (\%) on VisDA-2017 for unsupervised domain adaptation (ResNet-101). The best accuracy is indicated in bold and the second best one is underlined. * Reproduced by \cite{DSBN}}

\label{tab:VisDA}
\resizebox{\textwidth}{!}{%
\begin{tabular}{ccccccccccccc|c}
\hline
Method     & aero & bicycle       & bus  & car  & horse & knife & motor & person & plant & skate & train & truck & Avg  \\ 
\hline\hline
ResNet-101~\cite{ResNet1} & 72.3 & 6.1           & 63.4 & 91.7 & 52.7  & 7.9   & 80.1  & 5.6    & 90.1  & 18.5  & 78.1  & 25.9  & 49.4 \\
DANN~\cite{Ganin2015}    & 81.9 & 77.7          & 82.8 & 44.3 & 81.2  & 29.5  & 65.1  & 28.6   & 51.9  & 54.6  & 82.8  & 7.8   & 57.4 \\
DAN~\cite{DAN}        & 68.1 & 15.4          & 76.5 & \underline{87}   & 71.1  & 48.9  & 82.3  & 51.5   & 88.7  & 33.2  & \underline{88.9}  & 42.2  & 61.1 \\
MSTN*~\cite{MSTN}       & 89.3 & 49.5          & 74.3 & 67.6 & 90.1  & 16.6  & \textbf{93.6}  & 70.1   & 86.5  & 40.4  & 83.2  & 18.5  & 65.0 \\
JAN~\cite{Long2017}        & 75.7 & 18.7          & 82.3 & 86.3 & 70.2  & 56.9  & 80.5  & 53.8   & 92.5  & 32.2  & 84.5  & \underline{54.5}  & 65.7 \\
DM-ADA~\cite{Minghao2020}     & -    & -             & -    & -    & -     & -     & -     & -      & -     & -     & -     & -     & 75.6 \\
DMRL~\cite{Wu2020}       & -    & -             & -    & -    & -     & -     & -     & -      & -     & -     & -     & -     & 75.5 \\
MODEL~\cite{MODEL}      & 94.8 & 73.4          & 68.8 & 74.8 & 93.1  & \underline{95.4}  & 88.6  & \underline{84.7}   & 89.1  & 84.7  & 83.5  & 48.1  & 81.6 \\
STAR~\cite{STAR}       & 95   & 84            & \underline{84.6} & 73   & 91.6  & 91.8  & 85.9  & 78.4   & 94.4  & 84.7  & 87    & 42.2  & \underline{82.7} \\
CAN~\cite{CAN2019} &
  \textbf{97} &
  \underline{87.2} &
  82.5 &
  74.3 &
  \textbf{97.8} &
  \textbf{96.2} &
  90.8 &
  80.7 &
  \underline{96.6} &
  \textbf{96.3} &
  87.5 &
  \textbf{59.9} &
  \textbf{87.2} \\
\hline
FixBi (Ours) &
  \underline{96.1} &
  \textbf{87.8} &
  \textbf{90.5} &
  \textbf{90.3} &
  \underline{96.8} &
  95.3 &
  \underline{92.8} &
  \textbf{88.7} &
  \textbf{97.2} &
  \underline{94.2} &
  \textbf{90.9} &
  25.7 &
  \textbf{87.2} \\ \hline
\end{tabular}%
}
\end{table*}

\subsection{Comparison with the state-of-the-art methods}
We compare our method with the various state-of-the-art methods on three public benchmarks. The results of Office-31, Office-Home, and VisDA-2017 are reported in Tables \ref{tab:Office-31}, \ref{tab:Office-Home}, and \ref{tab:VisDA}, respectively. We use MSTN \cite{MSTN} as a baseline network. Note that our final accuracy is obtained by the sum of the softmax results of two network models. 

\textbf{Office-31.} Table~\ref{tab:Office-31} shows the comparative performance on the Office-31 dataset based on ResNet-50. The average accuracy of our method is 91.4\%, which outperforms the other methods such as SRDC~\cite{SRDC} and RSDA-MSTN~\cite{Gu2020}. Our method shows a significant performance improvement over the baseline MSTN \cite{MSTN} method in situations with very large domain shifts, e.g., A$\rightarrow$W, W$\rightarrow$A, A$\rightarrow$D, and D$\rightarrow$A tasks. In particular, compared to baseline, the task with the most improved accuracy was W$\rightarrow$A, achieving a performance improvement of 13.8\%. Compared with the mixup-based method DMRL \cite{Wu2020}, a large performance improvement is also observed. 

\textbf{Office-Home.} In Table~\ref{tab:Office-Home}, we compare our method with recent UDA methods on the Office-Home dataset based on ResNet-50. Our FixBi shows particularly strong performances in tasks with the large domain discrepancy between the real-world domain and the artificial domain, e.g., Rw$\rightarrow$Ar, Rw$\rightarrow$Cl, Rw$\rightarrow$Pr, and Cl$\rightarrow$Rw tasks. Our method has an average accuracy of 72.7\%, which outperforms the state-of-the-art results achieved by SRDC \cite{SRDC}. Note that our method achieves approximately 2\% better accuracy compared with RSDA-MSTN~\cite{Gu2020} on Office-Home.

\textbf{VisDA-2017.} Table~\ref{tab:VisDA} presents the classification accuracy for the VisDA-2017 dataset based on ResNet-101. Our FixBi achieves 22.2\% performance improvement on an average compared with the baseline method \cite{MSTN}. Above all, our method shows about 12\% better accuracy than other mixup-based methods~\cite{Wu2020, Minghao2020} and achieves comparable performance compared to the state-of-the-art methods.

\section{Conclusion}
In this paper, we proposed a FixBi algorithm that bridging the domain spaces to deal with the large domain discrepancy problem in an unsupervised domain adaptation scenario. Our main methodology is to construct an intermediate domain with different characteristics between the source domain and the target domain. We completed this through our fixed ratio-based mixup with different mixup-ratios, and further proposed bidirectional matching, self-penalization, and consistency regularization for efficient use of intermediate space.
Extensive ablation studies demonstrate the effectiveness of our proposed algorithm and experiments on the three standard benchmarks show that our proposed method achieves competitive performance to the state-of-the-art methods.

\textbf{Acknowledgement.} This work was partially supported by NRF-2020R1F1A1066049 and Technology Innovation Program (TIP-20000316) funded by the Ministry of Trade, Industry \& Energy (MOTIE, Korea).

{\small
\bibliographystyle{ieee}
\bibliography{egbib}
}

\end{document}